\documentclass[a4paper, 10pt, conference]{ieeeconf}      

\IEEEoverridecommandlockouts                              

\usepackage[pdftex] {graphicx}
\graphicspath{{figures/}}
\usepackage{hyperref}
\hypersetup{
	colorlinks=true,
	linkcolor=black,
	citecolor=black,
	filecolor=black,
	urlcolor=black,
}
\usepackage[T1]{fontenc}
\usepackage[utf8]{inputenc}
\usepackage{csquotes}
\usepackage[english]{babel}
\usepackage[export]{adjustbox}
\usepackage{caption}
\usepackage{subcaption}
\usepackage[colorinlistoftodos,backgroundcolor=orange!20,bordercolor=orange]{todonotes}
\usepackage{lipsum}
\usepackage[a4paper,left=.75in,right=.52in,top=1.46in,bottom=.785in]{geometry}
\usepackage{amsmath}
\usepackage{amssymb}
\usepackage{mathtools}
\usepackage{calc}
\usepackage{pgfplots}
\usepackage{siunitx}

\usepackage{placeins}

\usepackage[style=ieee,
doi=false,
url=false,
mincitenames=1,
maxcitenames=1,
minbibnames=6,
maxbibnames=6,
backend=biber]{biblatex}  
\title{\LARGE \bf Localization in Aerial Imagery with Grid Maps using LocGAN }

\author{
Haohao Hu$^1$, Junyi Zhu$^1$, Sascha Wirges$^2$ and Martin Lauer$^1$
\thanks{$^1$Authors are with Institute of Measurement and Control Systems, Karlsruhe Institute of Technology, Karlsruhe, Germany. {\tt\small \{haohao.hu, martin.lauer\}@kit.edu, junyi.zhu@gmx.de}}%
\thanks{$^2$Author is with FZI Research Center for Information Technology, Karlsruhe, Germany. {\tt\small wirges@fzi.de}}%
}

\begin{document}

\maketitle
\thispagestyle{empty}
\pagestyle{empty}
\begin{abstract}
In this work, we present LocGAN, our localization approach based on a geo-referenced aerial imagery and LiDAR grid maps.
Currently, most self-localization approaches relate the current sensor observations to a map generated from previously acquired data.
Unfortunately, this data is not always available and the generated maps are usually sensor setup specific.
Global Navigation Satellite Systems (GNSS) can overcome this problem.
However, they are not always reliable especially in urban areas due to multi-path and shadowing effects.
Since aerial imagery is usually available, we can use it as prior information.
To match aerial images with grid maps, we use conditional Generative Adversarial Networks (cGANs) which transform aerial images to the grid map domain.
The transformation between the predicted and measured grid map is estimated using a localization network (LocNet).
Given the geo-referenced aerial image transformation the vehicle pose can be estimated.
Evaluations performed on the data recorded in region Karlsruhe, Germany show that our LocGAN approach provides reliable global localization results.
\end{abstract}

\section{INTRODUCTION}
\label{sec:INTRODUCTION}
Accurate and robust localization is a foundation of autonomous vehicles.
Its main task is to estimate a transformation of the vehicle with respect to a globally referenced coordinate frame and the most common solution is GNSS.
However, due to multi-path and shadowing effects, GNSS are usually not reliable in urban areas.
To overcome this problem, most automated vehicles localize themselves in highly accurate maps, which are built based on onboard vision or range sensor data \cite{Sons2017SurroundViewML, Poggenhans2018PreciseLI}.
But these localization maps of the environment need to be created before driving automatically and their storage size may be very big, depending on which features are used.
Additionally, these maps are usually sensor setup specific and can only be used for similar sensor setups.
Besides that, such maps are usually build from multiple independent recording sessions \cite{Sons2018MultiDriveMapping}.
To archive a fusion of all sessions into one global frame, a geo-referenced aerial imagery is used in the approach proposed in \cite{Hu2019Georeference}.
To address this issue, we can also directly localize vehicles in a public available maps like the Open Street Map (OSM) or a aerial imagery, which are at most geo-referenced.
\begin{figure}
\centering
\includegraphics[width = \columnwidth]{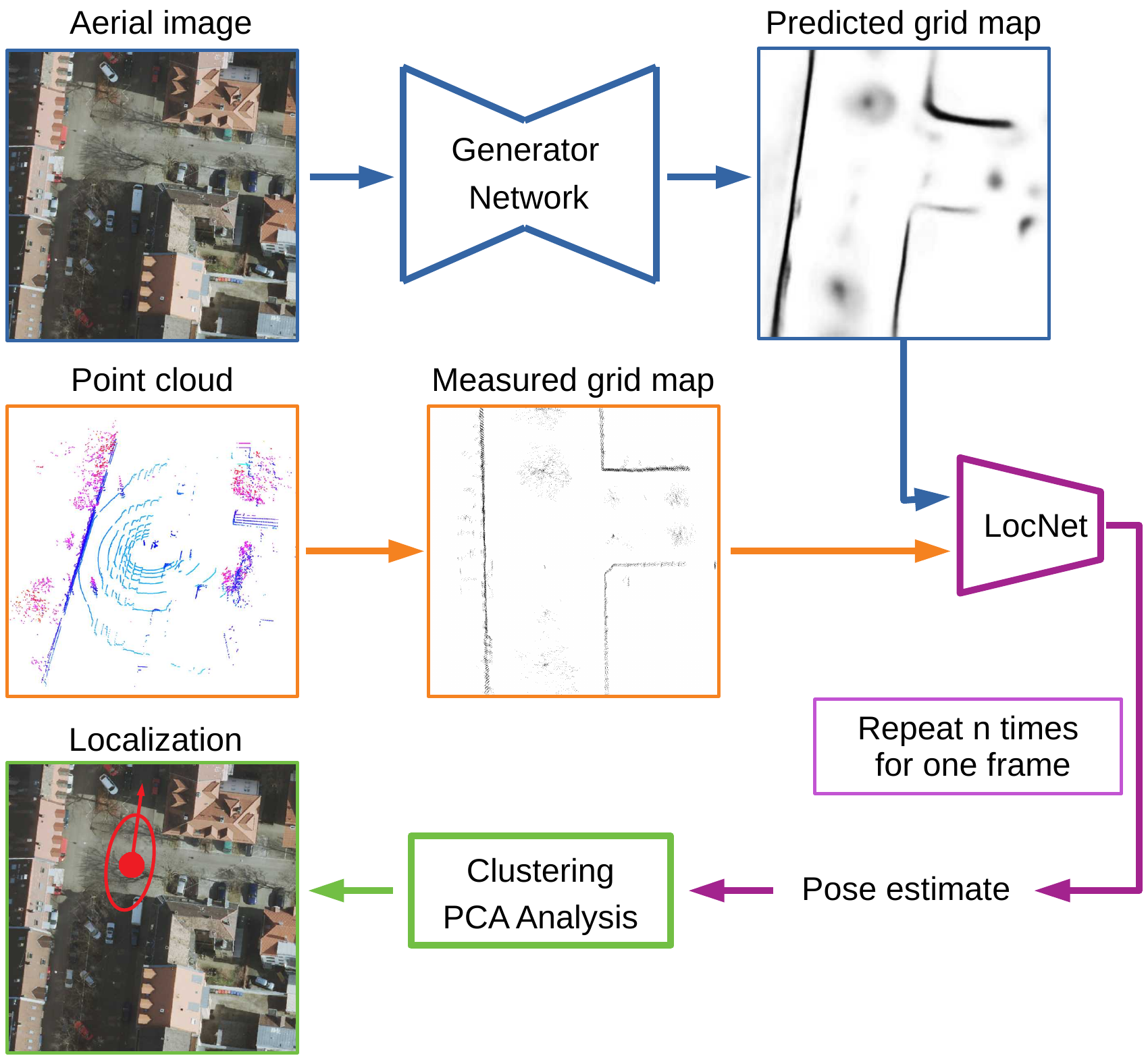}
\caption{
LocGAN.
First, we sample an image in aerial imagery w.r.t. a coarse GNSS localization.
Afterwards, we transform the image into the grid map domain using the generator part of cGAN. 
The LocNet can estimate the transformation between the predicted and measured grid map.
This process will be repeated n times per frame.
Finally, we apply clustering and PCA to estimate and analyze the final localization.
Aerial Imagery: \copyright Stadt Karlsruhe | Liegenschaftsamt.
}
\label{fig:motivation}
\end{figure}
In this work, we propose a novel approach to localize vehicles in a geo-referenced aerial imagery using grid maps.
We consider the problem of nowadays automated vehicle localization which is initialized by a coarse GNSS global localization.
The goal is to improve the final localization, so that the vehicles can confirm their own lane and their orientation in urban multi-lane scenarios.
As shown in Fig. \ref{fig:motivation}, we use the generator of a cGAN to transform aerial images into the grid map domain.
We then apply LocNet to estimate the transformation between the predicted and measured grid map.
With this transformation and the globally-referenced transformation of the aerial image we can determine a globally-referenced vehicle transformation based on aerial imagery.
This process will be repeated n times for each frame.
Finally, we improve the localization accuracy by using clustering based on multiple transformation estimates.
In experiments, we use the Principal Component Analysis (PCA) to analyze the localization uncertainty of our approach.
The main contribution of this work is the accurate vehicle localization in a geo-referenced aerial imagery using LiDAR grid maps and LocGAN, especially in the lateral direction, which gives rise to a reliable lane-level localization with a median error of $0.2$m.
The cGAN generator and LocNet can realize a successful matching between two different domains: the aerial imagery domain and the grid map domain.
\begin{figure*}
\includegraphics[width = \textwidth]{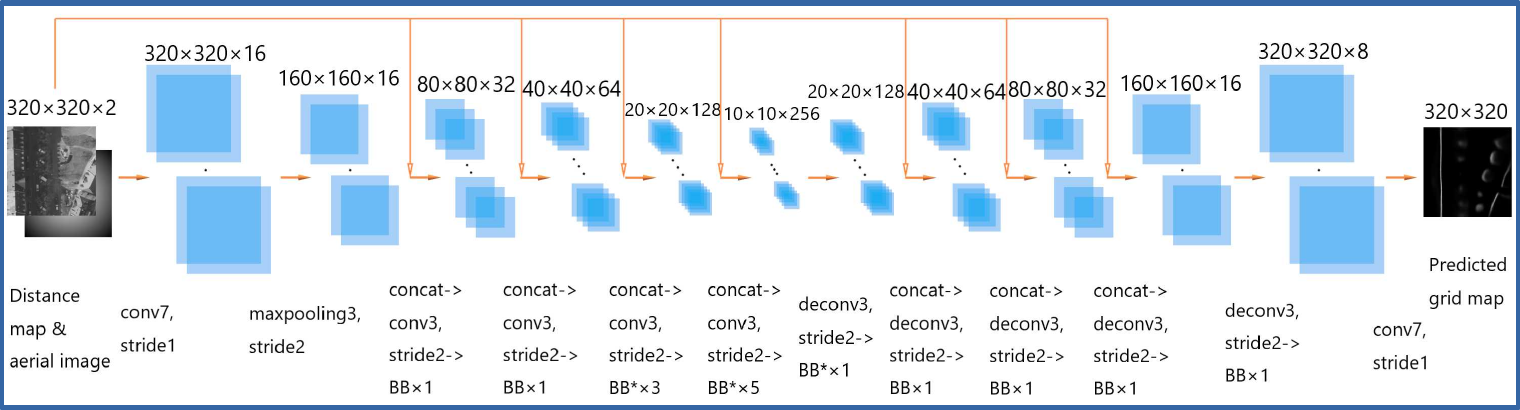}
\caption{
Generator.
By concatenation the input is resized to the feature map size.
BB is used as the abbreviation of the ResNet building block version 2 (V2).
The superscript * means a 1$\times$1 layer first reduces the number of feature maps to 64 then restores it after convolution with a 3$\times$3 layer.
As same as in the ResNet block V2, we use pre-activation (ReLU and batch normalization) for other layers except for the first one.
At last, we use a sigmoid to rescale the output into $(0, 1)$.
}
\label{fig:generator}
\end{figure*}
\section{ALGORITHM OVERVIEW}
\label{sec:ALGORITHM OVERVIEW}
The problem we want to solve in this work is vehicle self-localization in an aerial imagery using top-view grid maps.
We assume that the platform has one or more 3D LiDAR sensors scanning horizontally.
We use grid maps generated from point clouds as the detection input of our approach.  
Since grid maps and aerial imagery contain information from different domains, it is undesirable to match them directly.
To address this issue, We first use image-to-image translation techniques \cite{Isola2017ImageToImage} to train a generator, which maps aerial images into the grid map domain.  
We then consider the problem to estimate the transformation between the predicted and measured grid map as a flow estimation problem.
To accomplish this, we propose the LocNet based on the modified FlowNetS architecture.
Given the global transformation of each aerial image, the final localization is easy to estimate. 

\subsection{Network Design Concept}
\label{subsec:Network Design Concept}
The core of our approach is mapping aerial images to grid maps.
Since grid maps present the nearest outline of obstacles and aerial images contains more details, the mapping process can be regarded as a feature extraction process.
It does per-pixel prediction to estimate if the pixel belongs to the nearest outline of obstacles.
In the last few years, several approaches like Fully Convolutional Networks (FCNs) \cite{Long2015FCNs} and other autoencoder style Convolutional Neural Networks (CNNs) have been widely studied and achieve good results. 
But a naive FCN or other autoencoder style CNN can not accomplish the mapping issue in our application directly due to the following reasons:
1. They use a low-level objective, which is minimized by averaging all plausible outputs, causes blurring \cite{Isola2017ImageToImage}.
2. Since we use grid maps as ground truth, the measurement noise leads the network to do an ambiguous prediction within a local region rather than explicit prediction in a specific pixel.
3. Because the receptive fields of adjacent pixels overlap significantly, the sparse grid map structure will introduce conflicts in backpropagation and cause an ambiguous prediction.
Thus, it is desirable to use a high-level objective which owns an abstract comprehension of the whole predicted grid map rather than only being sensitive to single pixels.
As a result we make use of cGAN in this work, whose generator predicts grid maps from aerial images and discriminator provides a high-level objective.

\subsection{Implementation Details}
\label{subsec:Implementation Details}
Generative Adversarial Nets (GANs) \cite{Goodfellow2014GANs} consist of a generator $G$ and a discriminator $D$.
The generator learns a mapping from a random noise vector $z$ to data space $X$ as $\mathrm{G}(z; \theta_\mathrm{g})$.
The discriminator learns to maximize the probability of assigning the correct label to both real data samples $\mathrm{D}(x; \theta_\mathrm{d})$ and samples from the generator $\mathrm{D}(\mathrm{G}(z; \theta_\mathrm{g}); \theta_\mathrm{d})$.
GAN can be extended to cGAN if the generator and discriminator are conditioned on an extra information $y$ \cite{MirzaO14CGANs}.
In this work, aerial images perform the conditioning, which contains information $x$ should satisfy.
Since we prefer a deterministic predicted grid map rather than diverse output, so that the mapping function is $\mathrm{G}(y)$ rather than $\mathrm{G}(z|y)$.
For simplicity we still call the mapping $y \mapsto x$ a generator.

\subsubsection{Generator}
\label{subsubsec:Generator}
The generator is the pivotal part of our whole model.
In order to reduce compute costs, the input and output of the generator are 320 by 320 images.
They cover a $60$m by $60$m region in the real world and one pixel represents a $0.2$m by $0.2$m scope. 
%
%
Since our goal is determining on which lane the vehicle is driving, i.e. the localization error in lateral direction should be less than $1.5$m.
%
%
The predicted obstacle in grid maps should have high-precision..
To recover the outline of obstacles especially its precise position in images, we concatenate aerial images to feature maps during downsampling and upsampling.
Since the occlusion is hard to be learned by an CNN due to the weight sharing, we add a distance map as an additional channel to the input.
Every pixel in the distance map explicitly indicates how far it is from the aerial image center where the vehicle is located.
We also use ResNet units \cite{He2016ResNet} to make the network go deeper to extract high-level features, see Fig. \ref{fig:generator}.

\subsubsection{Adversarial Discriminator and LocNet}
\label{subsubsec:Discriminator and LocNet}
\begin{figure}[!h]
\begin{subfigure}[c]{\columnwidth}
\includegraphics[width=\columnwidth]{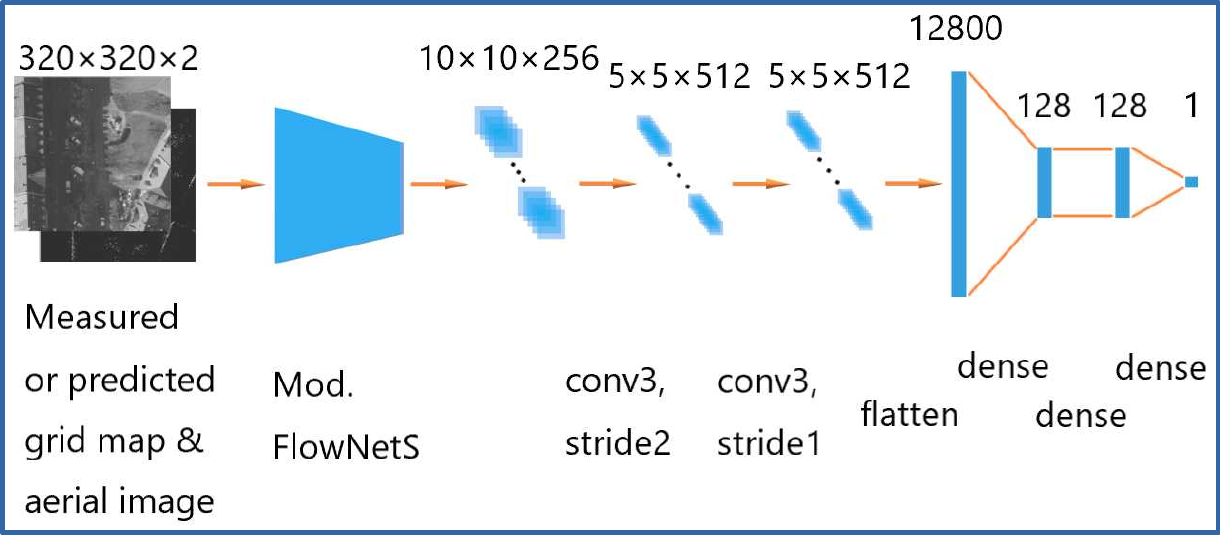} 
\subcaption{Discriminator}
\label{fig:flownet:a}
\end{subfigure}
\begin{subfigure}[c]{\columnwidth}
\includegraphics[width=\columnwidth]{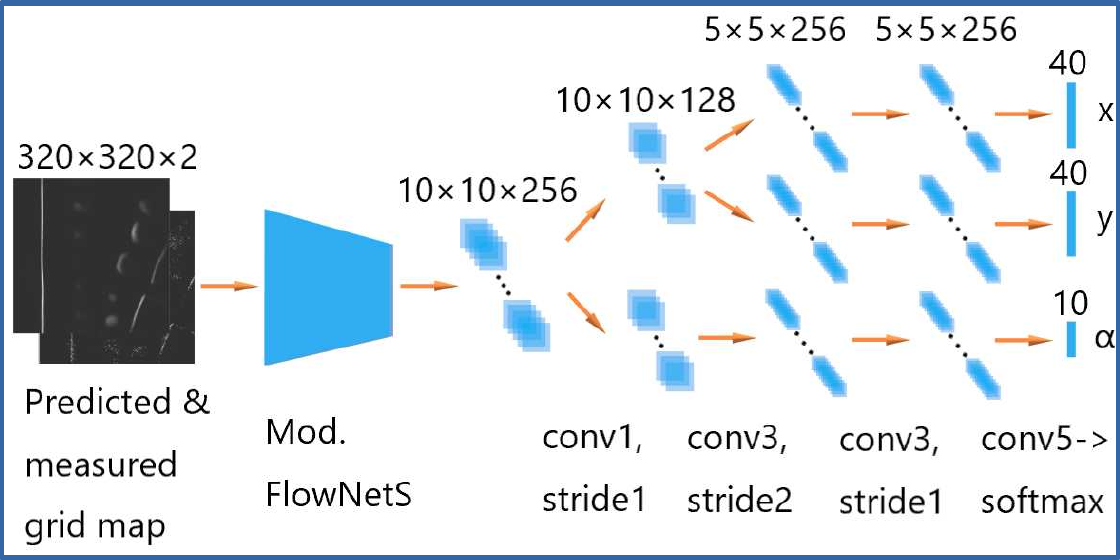} 
\subcaption{LocNet}
\label{fig:flownet:b}
\end{subfigure}
\begin{subfigure}[c]{\columnwidth}
\includegraphics[width=\columnwidth]{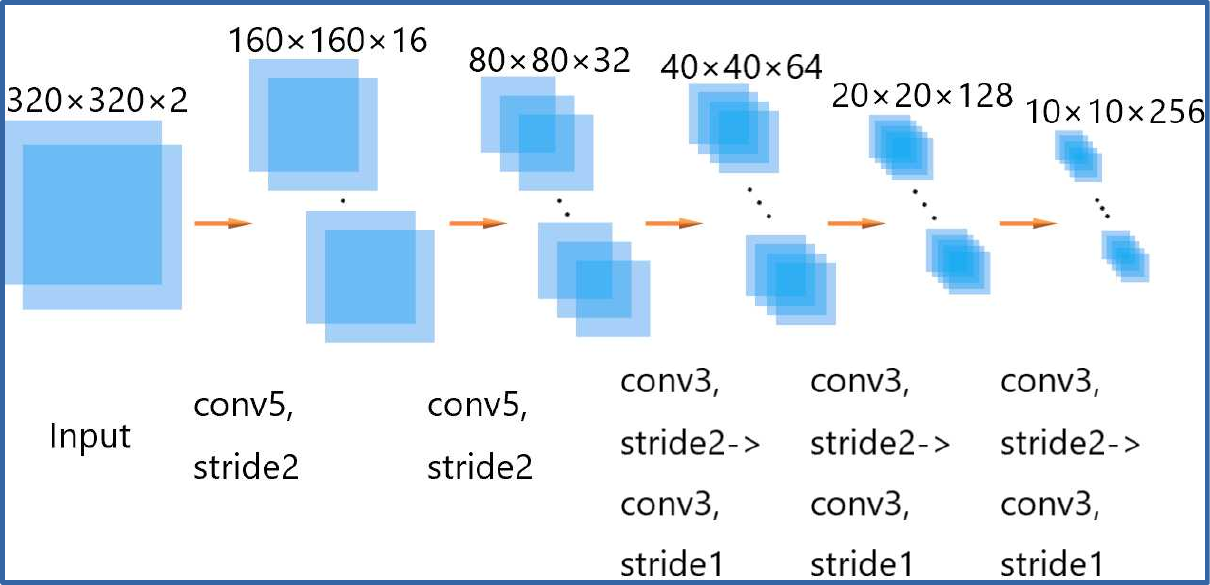} 
\subcaption{Modified FlowNetS}
\label{fig:flownet:c}
\end{subfigure}
\caption{
Discriminator and LocNet.
In Fig. \ref{fig:flownet:a} and \ref{fig:flownet:b}, all layers use post-activation (batch normalization and leaky-ReLU) except the last one which removes batch normalization.
In Fig. \ref{fig:flownet:c}, the details of modified Flownets are presented.
}
\label{fig:flownet}
\end{figure}
The predicted grid map is first processed using the pixel-wise objective presented in Eq. \ref{eq:pixel_loss}, where $p$ and $g$ are the predicted and measured grid map.
\begin{equation}
\mathcal{L}_\mathrm{pixel} = \sum\nolimits_\mathrm{ij} -(g_\mathrm{ij}\log(p_\mathrm{ij}) + (1-g_\mathrm{ij})\log(1-p_\mathrm{ij}))
\label{eq:pixel_loss}
\end{equation}
But the pixel-wise objective is not sufficient to get a refined grid map. So we further train the generator by means of the discriminator.
The discriminator architecture follows the FlowNetS \cite{Dosovitskiy2015FlowNet}, see Fig \ref{fig:flownet:c}.
Instead of upsampling we use fully connected layers to output one value, see Fig.\ref{fig:flownet:a}, which represents the estimated probability.
The discriminator optimize the generator in two aspects: the predicted grid map should look like a measured grid map and precisely recover its position in the aerial image.
Eq. \ref{eq:GAN_loss} represents the objective, where $\mathrm{G}$ tries to minimize this objective against $D$ that tries to maximize it:
\begin{align}
\mathcal{L}_\mathrm{cGAN} & = E_\mathrm{x \sim X_\mathrm{real}}[\log(\mathrm{D}(x, y))] \hspace{5pt} + \nonumber \\
                           & \hspace{13pt} E_\mathrm{x \sim X_\mathrm{predicted}}[\log(1 - \mathrm{D}(\mathrm{G}(y), y)]
\label{eq:GAN_loss}
\end{align}
For localization, we also follow FlowNetS, and replace the upsampling part with three CNN branches.
The output $x, y, \alpha$ indicate the probability distributions of the transformation between the predicted and measured grid map, see Fig. \ref{fig:flownet:b}. 
We assume that the GNSS localization has at most $10$m error in the $x, y$ direction, and $10^\circ$ orientation error.
Furthermore, we discretize the x- and y-direction with a step size of $0.5$m in the range of $(-10\mathrm{m}, 10\mathrm{m})$ as well as the orientation $\alpha$ with $2^\circ$ in the range $(-10^\circ, 10^\circ)$.
The LocNet is optimized using the objective, where $g$ is the ground truth:
\begin{equation}
\mathcal{L}_{LocNet} = \sum\limits_\mathrm{ij} - g_\mathrm{ij}\log(\delta_\mathrm{ij}), \mathrm{i\in\{x, y, \alpha\}
\label{eq:LocNet_loss}}
\end{equation}

\subsection{Training}
\label{subsec:Training}
Since the roof color is a regional characteristic, considering the generalization we train the network with grayscale aerial images. 
An important issue with training is the imbalanced dataset. 
For instance, vehicles usually drive along the road, so the generator is prone to place the predicted obstacle outline parallel to the ego vehicle even by the lane changing which is definitely not the case.
To address this issue, we apply data augmentation techniques including flipping, scaling, varying brightness, rotation and translation to the aerial images.
To avoid a padded background, we perform an central crop to the processed images after rotation or translation, keep the part of the original image, then resize it to the original image size.
We train the generator using pixel-wise objective to learn a rough model distribution and then perform grid map refinement by means of the discriminator.
Since a model collapses easily happens in practice, we still keep the pixel-wise objective to stabilize the generator.
Finally, we train the LocNet to estimate the transformation between the sampled aerial image and the true vehicle pose. 
In training we use soft class labels due to the dependency between neighboring classes.

\subsection{Prediction}
\label{subsec:Prediction}
We predict $n$ grid maps for each frame, then estimate a pose difference $\Delta p$ for each predicted grid map.
The list of these poses is $\Delta P = \{\Delta p_\mathrm{i}; \mathrm{i} \in \mathbb{N} , 0 \le \mathrm{i} < n\}$.
In order to reject outliers, we cluster the poses using the Euclidean distance metric and only keep the cluster having most points.
We apply the PCA to analyze, how the filtered poses are distributed around the true vehicle position in the lateral and longitudinal direction.
Finally, the final vehicle localization is estimated using the mean of the filtered poses. 
\section{EXPERIMENTAL EVALUATION}
\label{sec:EXPERIMENTAL EVALUATION}
%
%
%
%
%

\subsection{Dataset}
\label{subsec:Dataset}
\begin{figure}
\centering
\includegraphics[width = \columnwidth]{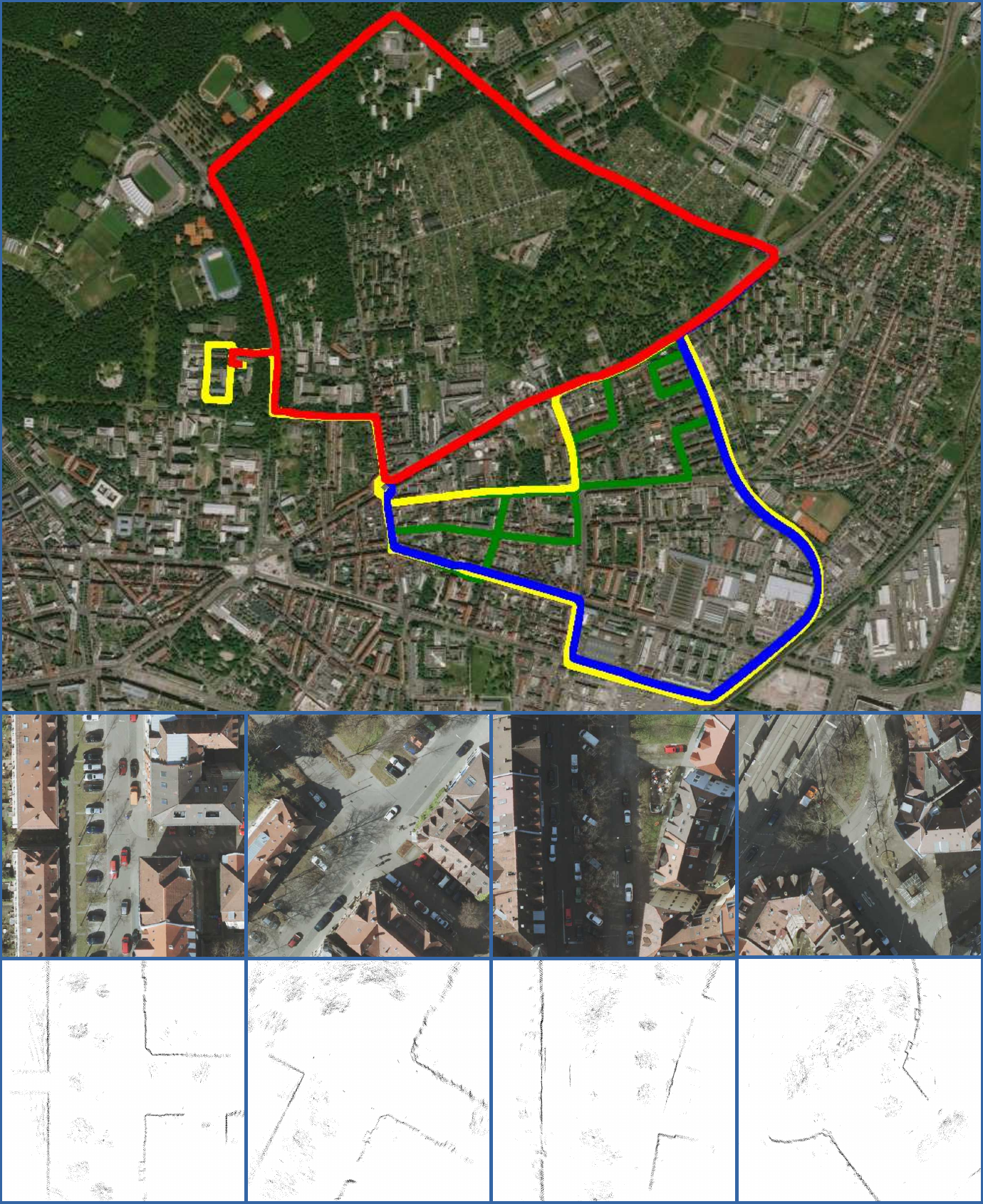}
\caption{
Upper part: Four passes recorded in region Karlsruhe, Germany, with a total driving distance of $24.3$km (Aerial Imagery: \copyright Bing Aerial Imagery).
Middle and lower part: aerial image examples cropped from the aerial imagery database and the corresponding grid maps (Aerial Imagery: \copyright Stadt Karlsruhe | Liegenschaftsamt).
}
\label{fig:dataset}
\end{figure}
For the training and evaluation, we recorded data with our experimental vehicle \emph{BerthaOne} \cite{Tacs2018Making}.
The dataset comprises data recorded from four Velodyne VLP16 LiDARs mounted flat on the roof, three BlackFly PGE-50S5M cameras behind the front- and rear windshield, and a Ublox C94-M8P GNSS receiver.
All sensors are jointly calibrated using approaches proposed in \cite{Strauss2014Calibration, Kummerle2018Calibration}.
The data contains four passes mapped and geo-referenced using approaches proposed in \cite{Sons2018Mapping, Hu2019Georeference}.
All passes with corresponding grid maps and aerial images are shown in Fig. \ref{fig:dataset}.
Since long-term obstacles are more useful for localization and most dynamic obstacles are below $3$m, we only use points at a hight over $3$m for the grid map generation \cite{Wirges2018GridMap}.
Aerial images are sampled every $3$m along the trajectory to reduce the data correlation.
Finally, we got 3590 aligned aerial image and grid map pairs, which are divided into training ($80\%$), testing ($10\%$) and evaluation data ($10\%$).
During the evaluation, we sample 100 aerial images for each frame around the vehicle true pose according to a uniform distribution in a range of $(-10\mathrm{m}, 10\mathrm{m})$ for the $x, y$ direction, and $(-10^\circ, 10^\circ)$ for orientation.

\subsection{Predicted Grid Maps}
\label{subsec:Predicted Grid maps}
\begin{figure}[!h]
\begin{subfigure}{0.49\columnwidth}
\includegraphics[width=\columnwidth]{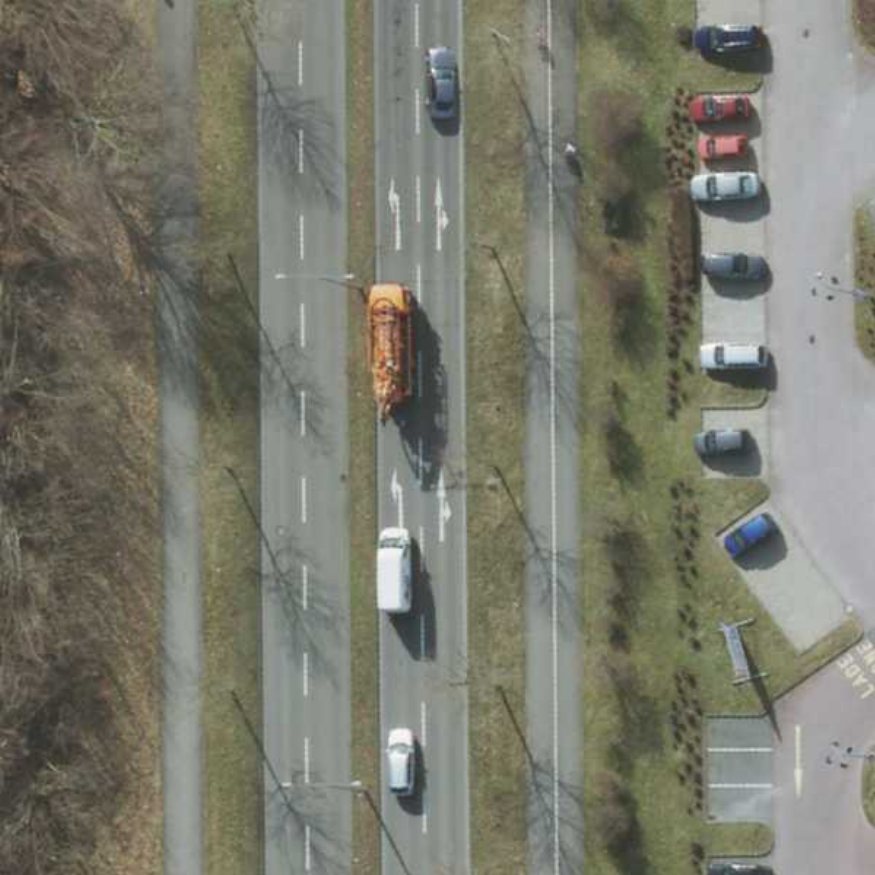} 
\subcaption{}
\label{fig:predicted_grid_maps:a}
\end{subfigure}
\hfill
\begin{subfigure}{0.49\columnwidth}
\includegraphics[width=\columnwidth]{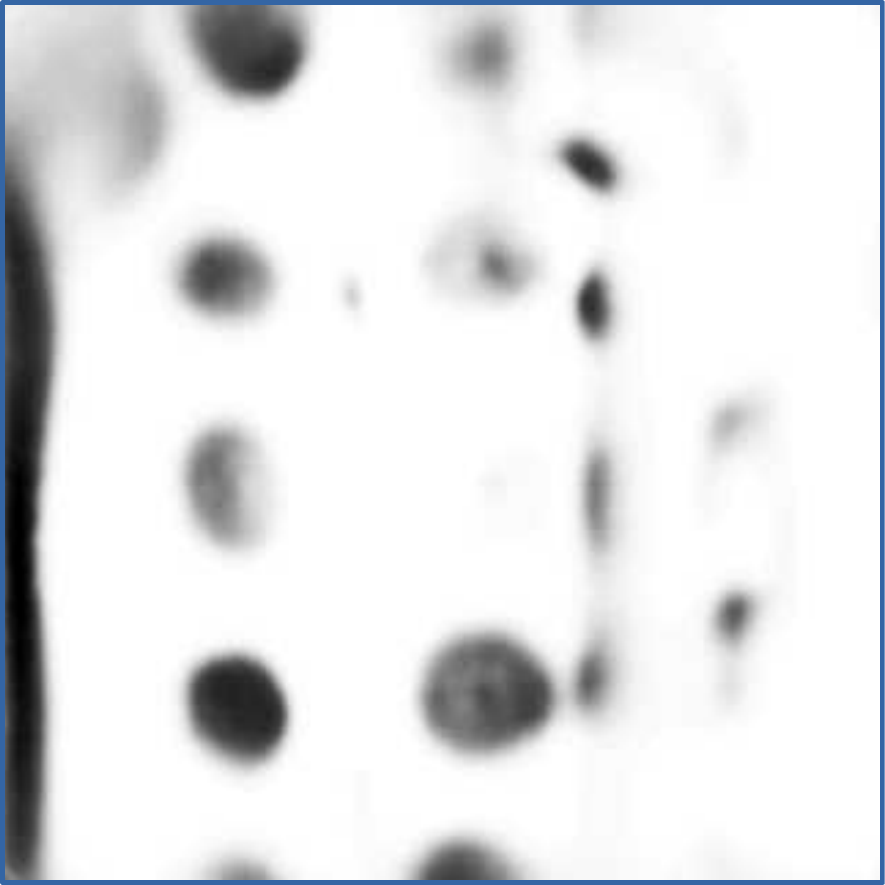} 
\subcaption{}
\label{fig:predicted_grid_maps:b}
\end{subfigure}
\begin{subfigure}{0.49\columnwidth}
\includegraphics[width=\columnwidth]{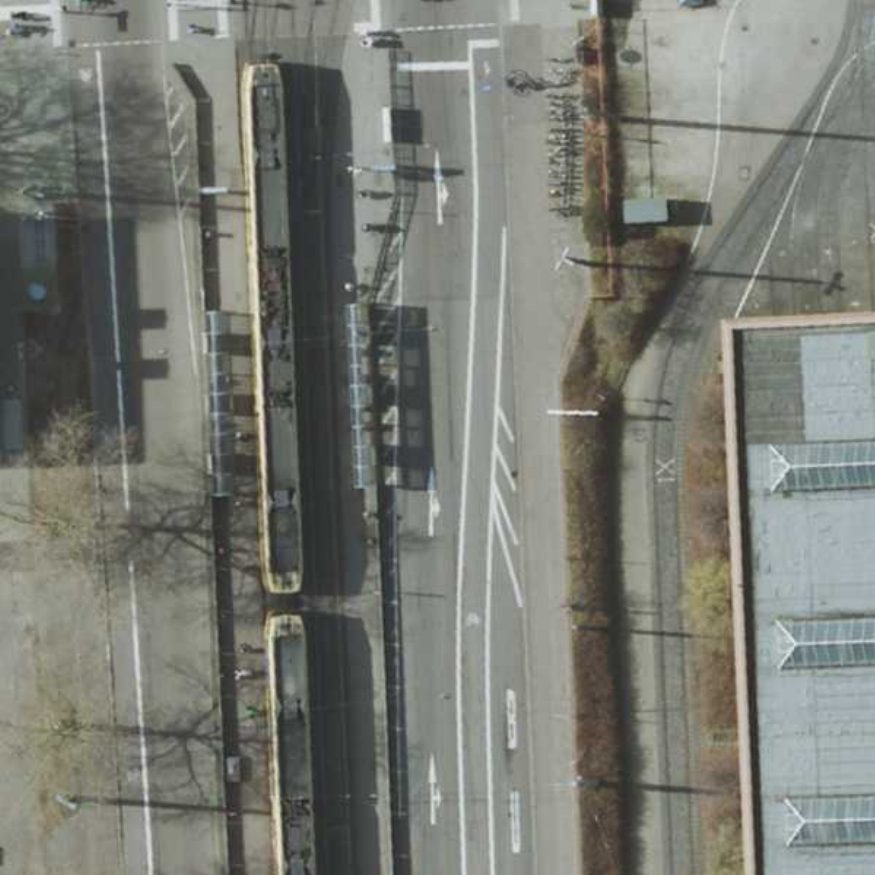} 
\subcaption{}
\label{fig:predicted_grid_maps:c}
\end{subfigure}
\hfill
\begin{subfigure}{0.49\columnwidth}
\includegraphics[width=\columnwidth]{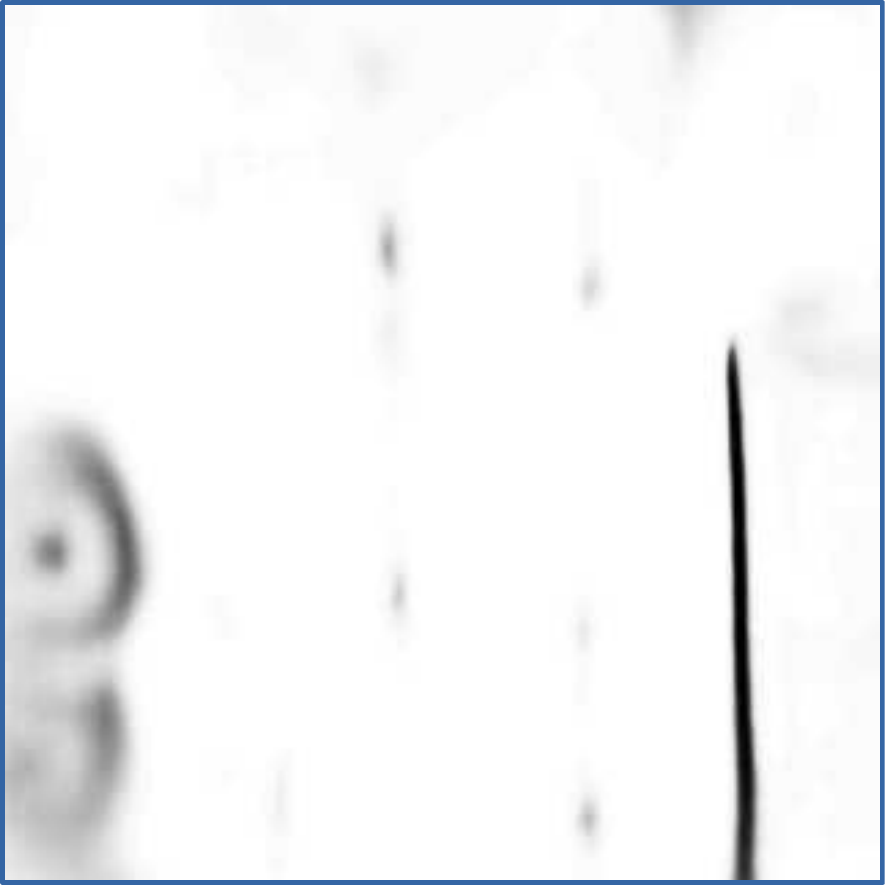} 
\subcaption{}
\label{fig:predicted_grid_maps:d}
\end{subfigure}
\begin{subfigure}{0.49\columnwidth}
\includegraphics[width=\columnwidth]{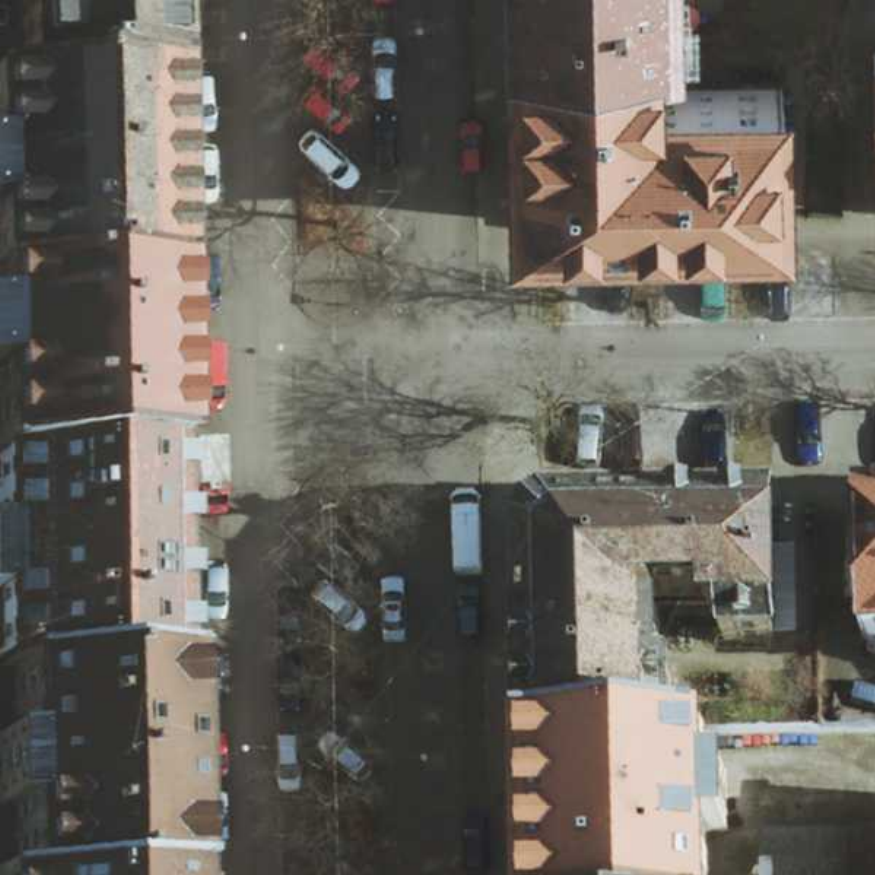} 
\subcaption{}
\label{fig:predicted_grid_maps:e}
\end{subfigure}
\hfill
\begin{subfigure}{0.49\columnwidth}
\includegraphics[width=\columnwidth]{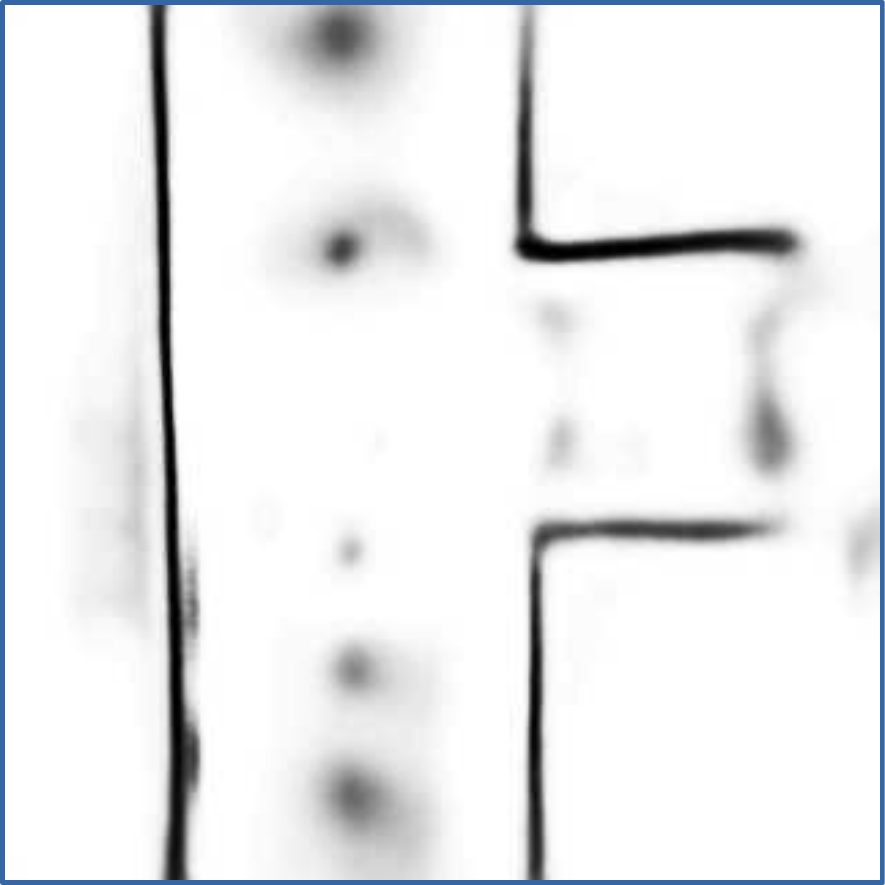} 
\subcaption{}
\label{fig:predicted_grid_maps:f}
\end{subfigure}
\caption{
Predicted grid maps from the cGAN generator.
Left images show the sampled aerial images from the aerial imagery and right images are the predicted grid maps.
Aerial Imagery: \copyright Stadt Karlsruhe | Liegenschaftsamt.
}
\label{fig:predicted_grid_maps}
\end{figure}
In Fig. \ref{fig:predicted_grid_maps}, the from our cGAN generator predicted grid maps for three typical scenarios are shown.
In the scenario shown in Fig. \ref{fig:predicted_grid_maps:e}, our generator predicts obstacle outlines in consideration of the occlusive relationship and it is robust against shadows, see Fig. \ref{fig:predicted_grid_maps:f}.
In Figs. \ref{fig:predicted_grid_maps:b} and \ref{fig:predicted_grid_maps:d}, additionally, the tree branch geometry is predicted accurately in the grid map, which helps the localization a lot in scenarios with few buildings.
As shown in all these scenarios, our generator has learned to ignore irrelevant features like vehicles and trams, which is a big benefit in comparison to other common approaches like edge detector and ray tracing.
All in all, our generator provides accurate grid map prediction based on aerial images for diverse scenario types.

\subsection{LocNet Prediction Analysis}
\label{subsec:LocNet Prediction Analysis}
\begin{figure}[!h]
\begin{subfigure}{0.49\columnwidth}
\includegraphics[width=\columnwidth]{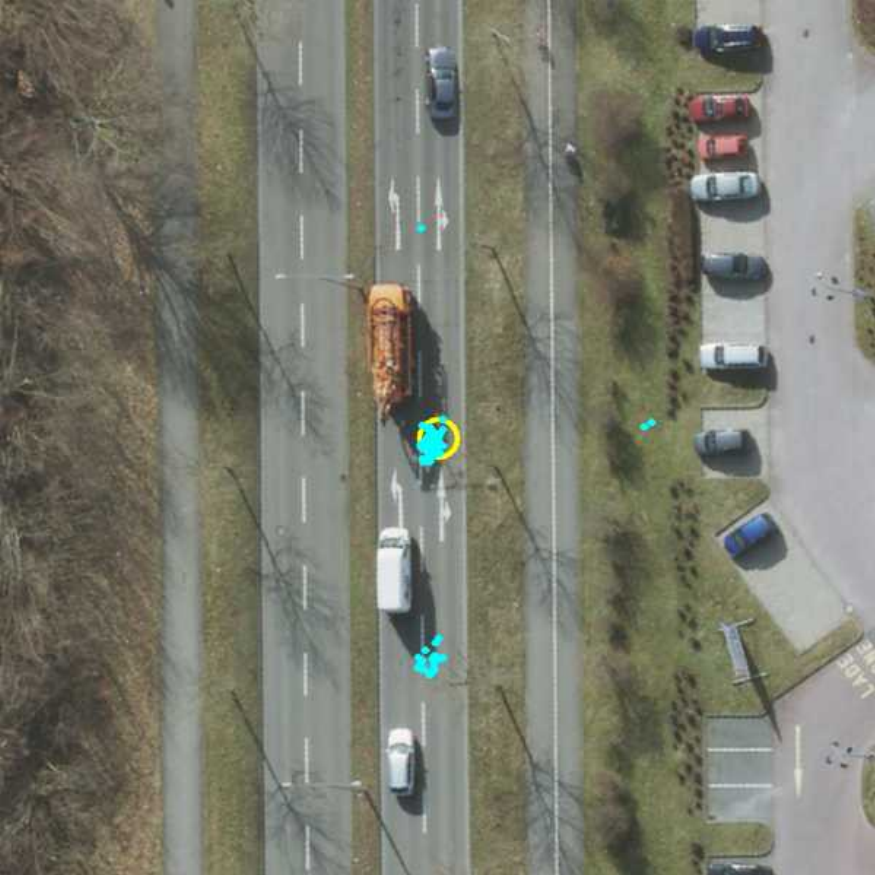} 
\subcaption{}
\label{fig:prediction_analysis:a}
\end{subfigure}
\hfill
\begin{subfigure}{0.49\columnwidth}
\includegraphics[width=\columnwidth]{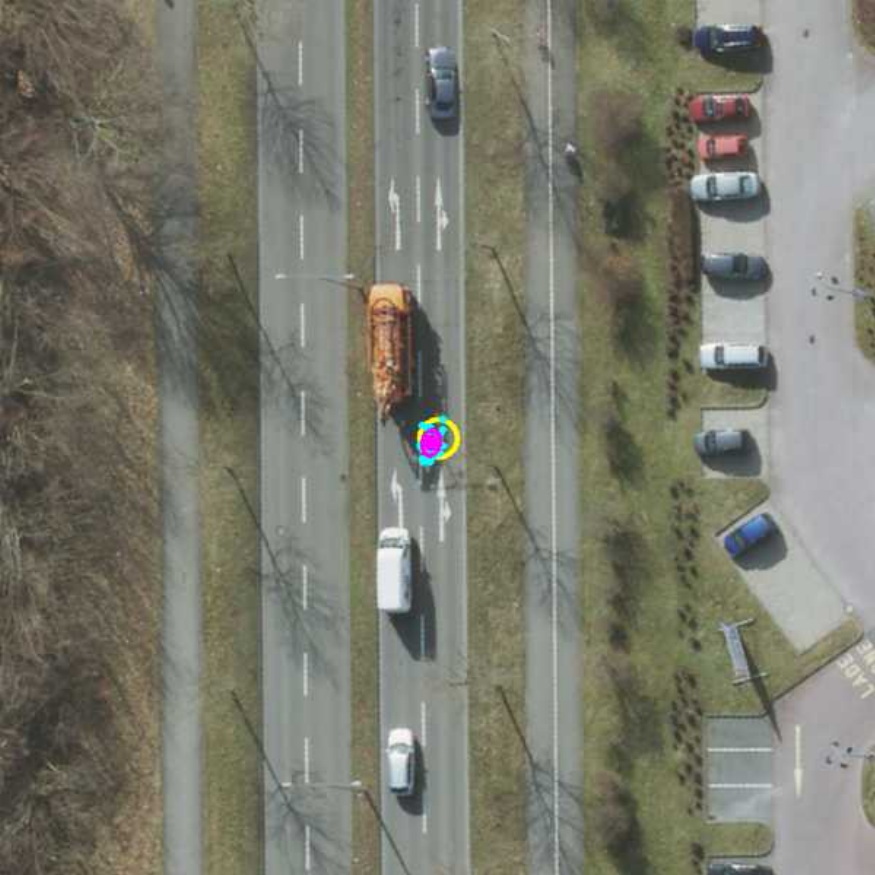} 
\subcaption{}
\label{fig:prediction_analysis:b}
\end{subfigure}
\begin{subfigure}{0.49\columnwidth}
\includegraphics[width=\columnwidth]{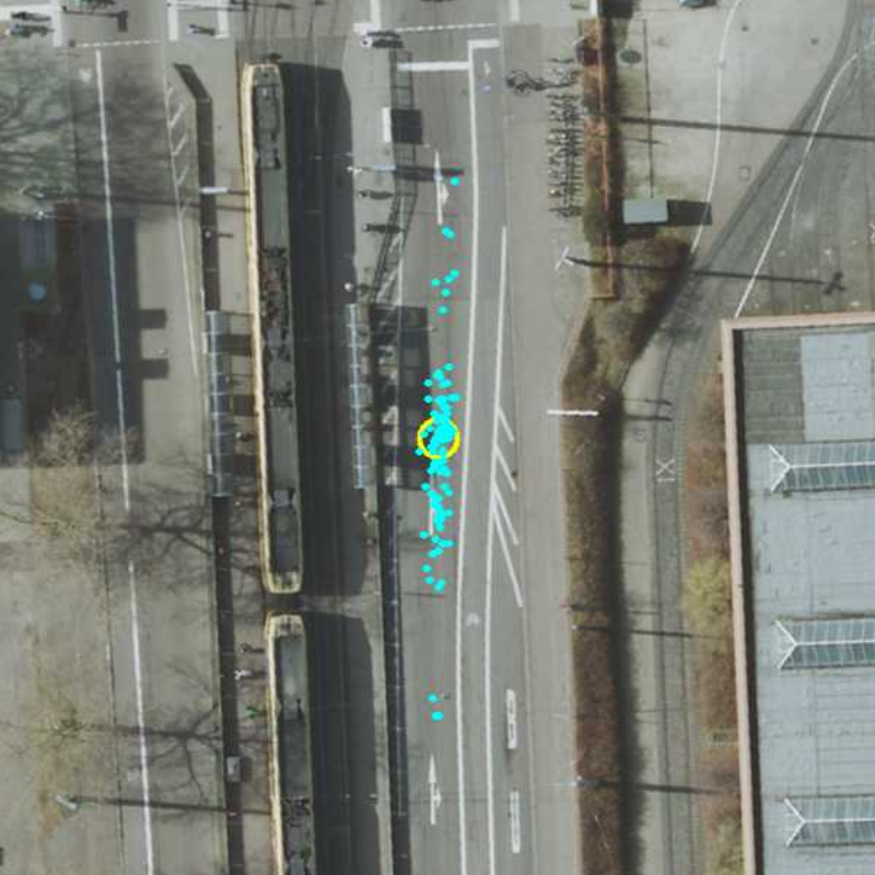} 
\subcaption{}
\label{fig:prediction_analysis:c}
\end{subfigure}
\hfill
\begin{subfigure}{0.49\columnwidth}
\includegraphics[width=\columnwidth]{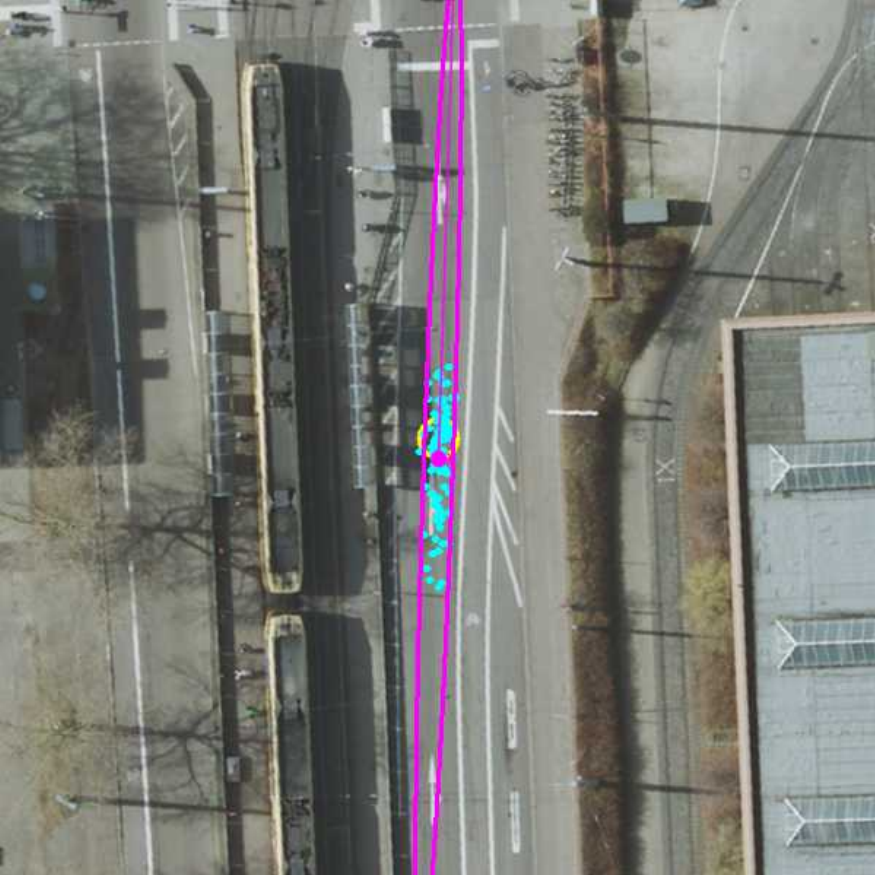} 
\subcaption{}
\label{fig:prediction_analysis:d}
\end{subfigure}
\begin{subfigure}{0.49\columnwidth}
\includegraphics[width=\columnwidth]{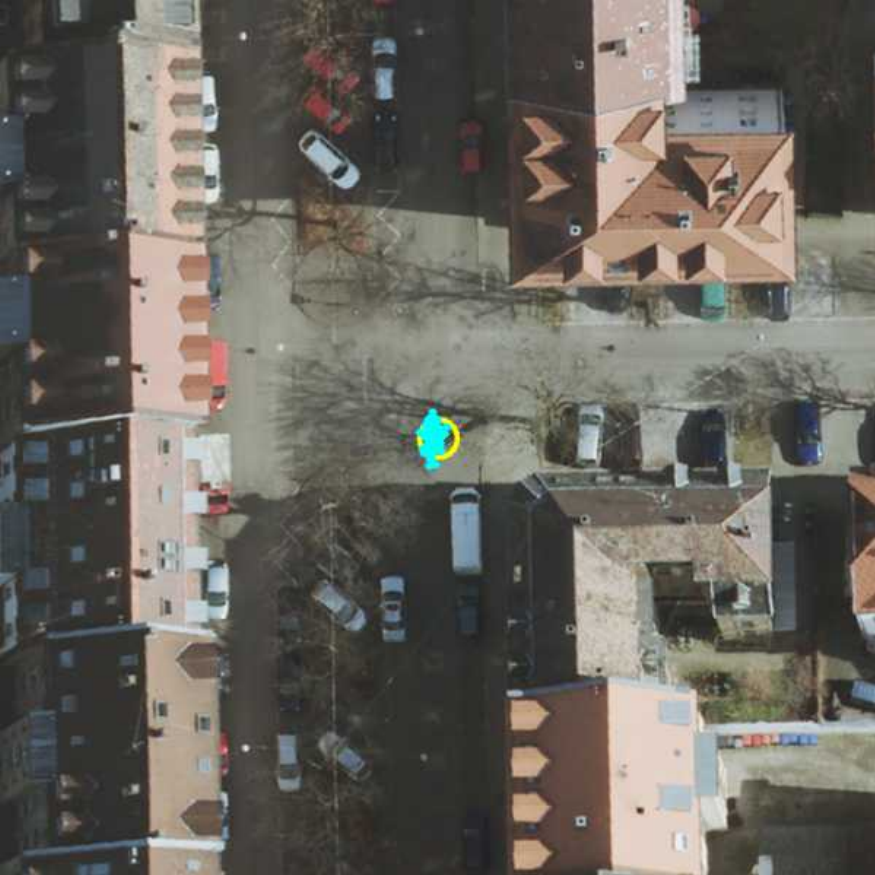} 
\subcaption{}
\label{fig:prediction_analysis:e}
\end{subfigure}
\hfill
\begin{subfigure}{0.49\columnwidth}
\includegraphics[width=\columnwidth]{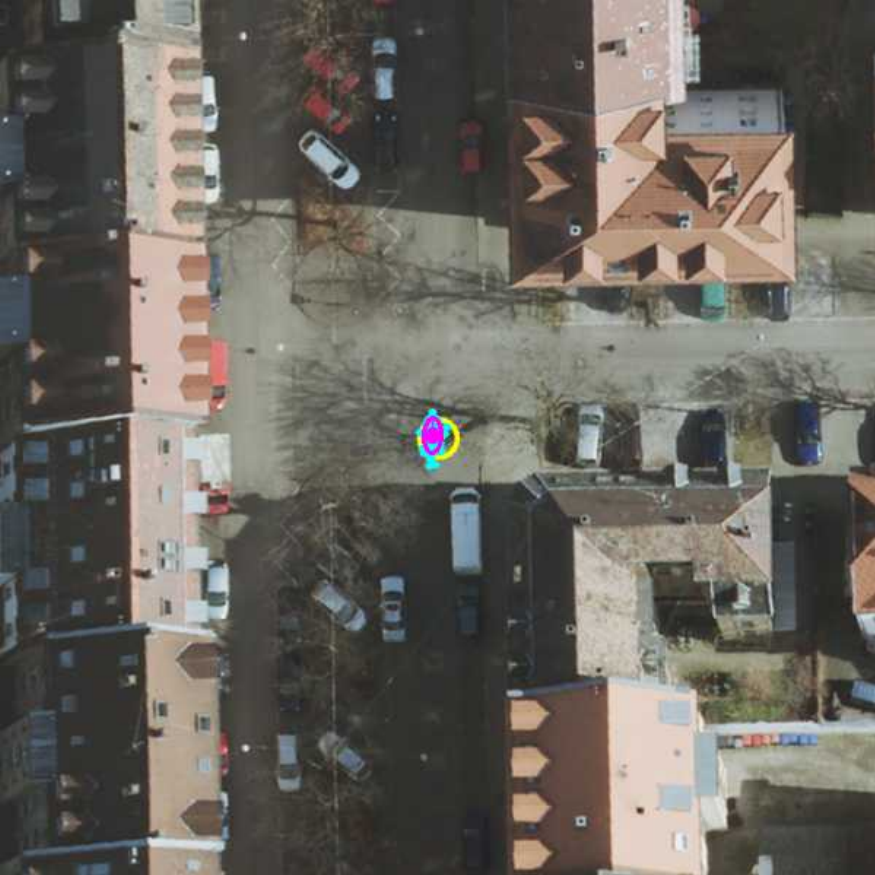} 
\subcaption{}
\label{fig:prediction_analysis:f}
\end{subfigure}
\caption{
LocNet prediction analysis with clustering and PCA.
The yellow circle shows the true vehicle position.
The left images show the localization predictions from LocNet with cyan points.
In the right images, the magenta ellipse presents the distribution of predictions filtered by clustering and the fat magenta point shows the final localization.
Aerial Imagery: \copyright Stadt Karlsruhe | Liegenschaftsamt.
}
\label{fig:prediction_analysis}
\end{figure}
Using scenarios shown in Fig. \ref{fig:prediction_analysis}, we analyze our localization predictions with clustering and PCA.
The scenario shown in Figs. \ref{fig:prediction_analysis:e} and \ref{fig:prediction_analysis:f} contains significant features in the lateral and longitudinal direction.
The localization predictions are close to each other and the final localization is precise and accurate.
The scenario shown in Figs. \ref{fig:prediction_analysis:c} and \ref{fig:prediction_analysis:d} contains less lateral features.
The lateral localization is more accurate than the longitudinal localization.
In the scenario shown in Figs. \ref{fig:prediction_analysis:a} and \ref{fig:prediction_analysis:b}, an accurate lateral localization is provided with the accurate prediction of trees in grid map.
However, several candidates are provided due to periodical signals in the longitudinal direction.
This issue is solved in this work by applying clustering and to solve it systematically, we can apply classical approaches like Iterative Closest Point (ICP) \cite{Rusinkiewicz2001ICP} using LocNet candidates as initialization to make the alignment confident.

\subsection{Localization Accuracy}
\label{subsec:Localization Accuracy}
\begin{figure}
\centering
\includegraphics[width = \columnwidth]{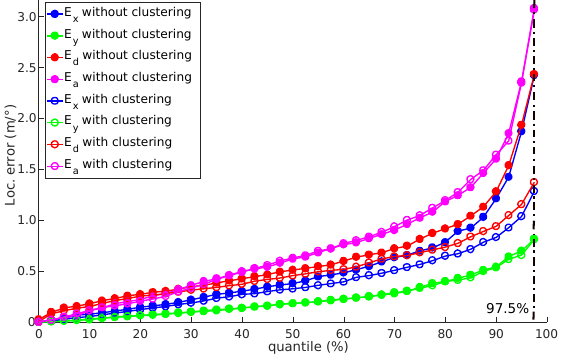}
\caption{
Loc. error analysis w.r.t. longitudinal direction ($E_\mathrm{x}$), lateral direction ($E_\mathrm{y}$), Euclidean distance ($E_\mathrm{d}$) and orientation ($E_\mathrm{a}$) with and without clustering.
The x-axis shows the quantile and the y-axis presents the localization error.
}
\label{fig:loc_accuracy}
\end{figure}
To evaluate our approach quantitatively, we analyze the localization accuracy considering: lateral, longitudinal, Euclidean distance and orientation.
The results are shown in Fig. \ref{fig:loc_accuracy}.
Comparing lines with filled circle (without clustering) and with circle (with clustering), the longitudinal localization is improved significantly using the clustering outlier rejecting.
Comparing blue lines ($E_\mathrm{x}$) with green lines ($E_\mathrm{y}$), we see the localization in lateral direction (median error $0.2$m) is more accurate than in longitudinal direction (median error $0.4$m), which is caused by poor longitudinal features in the dataset.
With this lateral accuracy, we get a lane-level localization, which is more important for autonomous vehicles compared to the longitudinal localization.
%
%

\section{RELATED WORK}
\label{sec:RELATED WORK}
Self-localization in a geo-referenced aerial imagery has the benefit that the localization result can be directly used in a geo-referenced LaneletMap \cite{Bender2014LaneletMap} for the trajectory planing and behavior generation of autonomous vehicles. 
Besides, the geo-referenced aerial imagery is usually public available.
As same as using a geo-referenced aerial imagery for localization, using the Open Street Map (OSM) has also the same benefits.
This topic is intensively researched in the last twenty years. 
This section reviews the state of the art of localization approaches using OSM or aerial imagery and make a comparing with our LocGAN approach.

\subsection{Localization using OSM}
\label{subsec:Localization using OSM}
In the approach proposed in \cite{Brubaker2013OSMRoadNetworkOdom}, the road network geometry extracted from OSM maps is used for autonomous vehicle localization with visual odometry.
Similarly, Ruchti et al. \cite{Ruchti2015OSMLiDAR} use the road network geometry for robot localization in urban areas by applying a learn-based approach to classify the point cloud grid into road or non-road.
The robot localization is provided through matching the classified grids with the extracted road network map.
Since the two above mentioned approaches exclusively use the road network geometry, they can only provide a road-level localization.
However, our approach provides a lane-level localization with a median error of $0.2$m in the lateral direction. 
In the appoach proposed in \cite{Hentschel2010OSMGeodata}, an OSM Geo-data map containing building edge information is used.
They detect the building edges using 3D LiDAR sensors and align the point cloud with the OSM Geo-data to estimate the vehicle pose.
Due to the poor features stored in the map, the localization accuracy can only achieve a mean error of $4.9$m.
In contrast, our generator predict building edges, tree geometries and other relevant features in grid maps.
The rich features lead our approach to achieve a median error of $0.5$m. 

\subsection{Localization using Aerial imagery}
\label{subsec:Localization using Aerial imagery}
The approach proposed in \cite{Kuemmerle2011SLAMAerialImagery} achieves the global consistency of Simultaneous Localization and Mapping (SLAM) by applying a public accessible aerial imagery as prior information.
This prior information can improve the global accuracy of SLAM significantly and the generated map is automatically geo-referenced.
However, their building structure extraction approach is not robust against non-building features like vehicles and trams.
In contrast, our generator is trained to ignore such dynamic obstacles and only predict long-term obstacles like buildings and trees in grid maps.
Through that, the localization robustness is improved significantly.  
Pink et al. \cite{Pink2009AerialImageryGlobalFeatureMap} extract road marking features from aerial imagery by applying a support vector machine classifier (SVM) to build a global feature map used for autonomous vehicle localization.
Since their approach exclusively uses road marking features, the vehicle can only be localized in intersections having many road markings.
Comparing to that, our approach can overcome this problem, because we use all of the long-term obstacles like buildings and trees in the environment for localization.
This makes the localization approach generally usable.

\section{CONCLUSION AND FUTURE WORK}
\label{sec:CONCLUSION AND FUTURE WORK}
In this work, we propose a vehicle self-localization approach LocGAN, which can localize in a geo-referenced aerial imagery using LiDAR grid maps.
As analyzed in Sec. \ref{sec:EXPERIMENTAL EVALUATION}, our LocGAN approach can provide reliable localization results in urban and suburb areas especially in the lateral direction with a median error of $0.2$m.
In comparison to another self-localization approaches using aerial imagery or OSM, our approach has improved the localization accuracy from a mean error of several meters to a median error of $0.5$m.
Our approach tries to localize vehicles using LiDAR grid maps and a geo-referenced aerial imagery, which however consist of information from two different domains.
This cross-domain extension make it possible to keep use of the geo-referenced aerial imagery as a localization map and add more different sensors like radar and camera into our LocGAN approach to improve our localization accuracy.
Using a Nvidia Geforce GTX TITAN X GPU, our network can process an aerial image within $5$ms.
In summary, using information from different domains, our LocGAN approach provides an accurate global localization compared to other approaches and it has the potential to provide an accurate global localization for autonomous vehicles in real time.
\section{Acknowledgment}
This research is accomplished within the project “UNICARagil”.
We acknowledge the financial support for the project by the Federal Ministry of Education and Research of Germany (BMBF). 
%
%
\printbibliography
\begin{figure*}
\centering
\includegraphics[width = \textwidth]{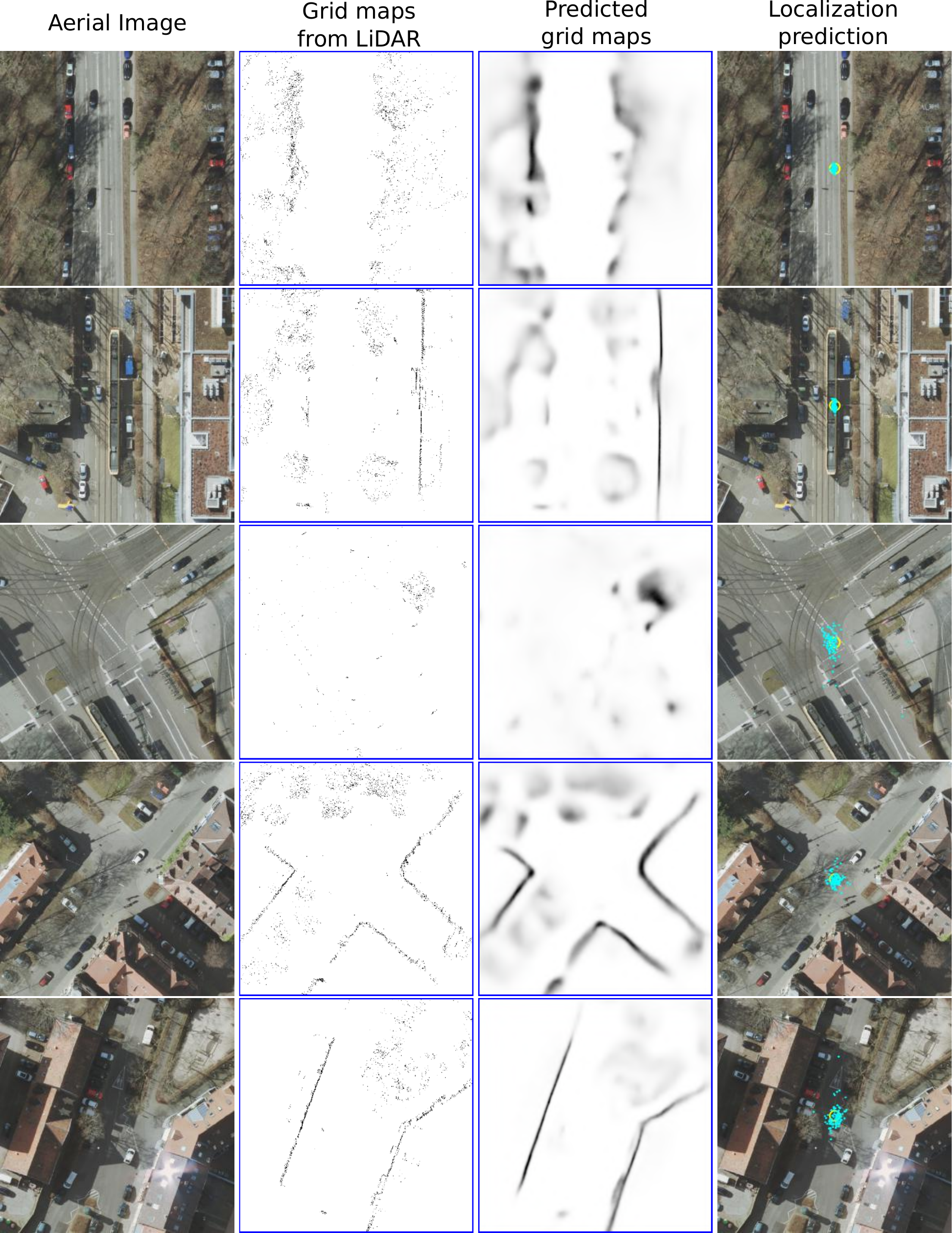}
\caption{
Examples of grid map prediction and localization prediction.
Aerial Imagery: \copyright Stadt Karlsruhe | Liegenschaftsamt.
}
\end{figure*}
\begin{figure*}
\centering
\includegraphics[width = \textwidth]{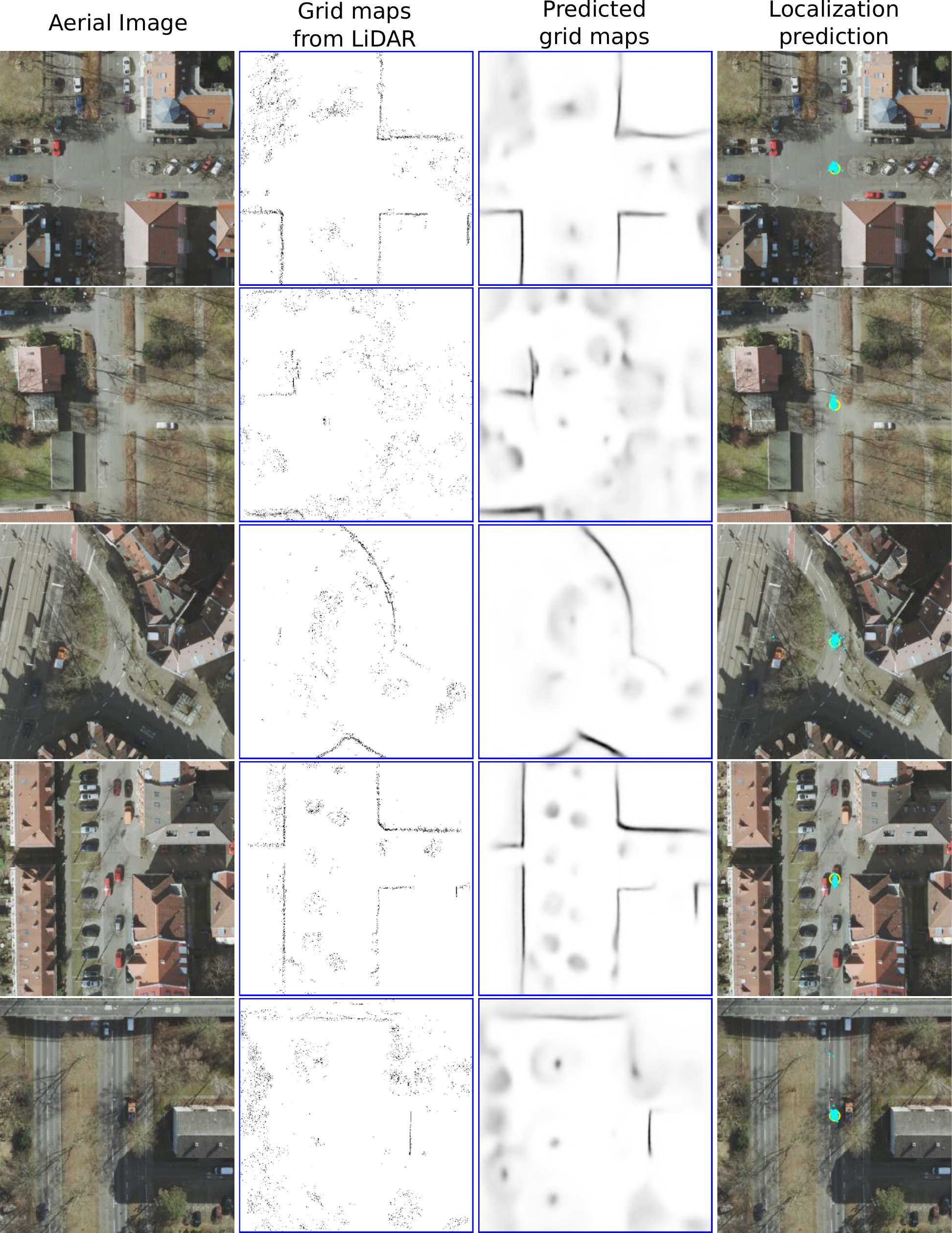}
\caption{
Examples of grid map prediction and localization prediction.
Aerial Imagery: \copyright Stadt Karlsruhe | Liegenschaftsamt.
}
\end{figure*}

\end{document}